\pdfoutput=1

\documentclass[runningheads]{llncs}
\usepackage[T1]{fontenc}
\usepackage{times}  % DO NOT CHANGE THIS
\usepackage{helvet}  % DO NOT CHANGE THIS
\usepackage{courier}  % DO NOT CHANGE THIS
\usepackage{graphicx} % DO NOT CHANGE THIS

\usepackage{times}
\usepackage{soul}
\usepackage{url}
\usepackage[hidelinks]{hyperref}
\usepackage[utf8]{inputenc}
\usepackage[small]{caption}
\usepackage{graphicx}
\usepackage{amsmath}
\usepackage{booktabs}
\usepackage[switch]{lineno}

\usepackage{algorithm2e}
\usepackage{algorithmic}

\usepackage{comment}
\usepackage{xcolor}

\usepackage{tikz}

\usepackage{listings}

\begin{document}

\title{Learning a Prior for Monte Carlo Search by Replaying Solutions to Combinatorial Problems}

\author{
    Tristan Cazenave
}
\authorrunning{Tristan Cazenave}
\titlerunning{Learning a Prior for Monte Carlo Search}
\institute{
LAMSADE, Université Paris Dauphine - PSL, CNRS, Paris, France
}

\maketitle

\begin{abstract}
Monte Carlo Search gives excellent results in multiple difficult combinatorial problems. Using a prior to perform non uniform playouts during the search improves a lot the results compared to uniform playouts. Handmade heuristics tailored to the combinatorial problem are often used as priors. We propose a method to automatically compute a prior. It uses statistics on solved problems. It is a simple and general method that incurs no computational cost at playout time and that brings large performance gains. The method is applied to three difficult combinatorial problems: Latin Square Completion, Kakuro, and Inverse RNA Folding.
\end{abstract}

\section{Introduction}

Monte Carlo Tree Search (MCTS) has been successfully applied to many games and problems \cite{BrownePWLCRTPSC2012}. It has superhuman performances in two player complete information games such as Go and Chess \cite{silver2017mastering}.

Nested Monte Carlo Search (NMCS) \cite{CazenaveIJCAI09} is an algorithm that works well for puzzles and combinatorial problems. It biases its playouts using lower level playouts. At level zero NMCS adopts a uniform random playout policy. Learning of playout strategies combined with NMCS has given good results on combinatorial problems \cite{RimmelEvo11}. Other applications of NMCS include Single Player General Game Playing \cite{Mehat2010}, Cooperative Pathfinding \cite{Bouzy13}, Software testing \cite{PouldingF14}, heuristic Model-Checking \cite{PouldingF15}, the Pancake problem \cite{Bouzy16}, Games \cite{CazenaveSST16}, the Inverse RNA Folding problem \cite{portela2018unexpectedly} and retrosynthesis \cite{roucairol2023retrosynthesis}.

Online learning of a playout policy in the context of nested searches has been further developed for puzzles and combinatorial problems with Nested Rollout Policy Adaptation (NRPA) \cite{Rosin2011}. NRPA has found new world records in Morpion Solitaire and crosswords puzzles. NRPA has been applied to multiple problems: the Traveling Salesman Problem
with Time Windows (TSPTW) \cite{cazenave2012tsptw,edelkamp2013algorithm}, 3D Packing with Object Orientation \cite{edelkamp2014monte}, the physical traveling salesman problem \cite{edelkamp2014solving}, the Multiple Sequence Alignment problem \cite{edelkamp2015monte} or Logistics \cite{edelkamp2016monte}. The principle of NRPA is to adapt the playout policy so as to reinforce the best sequence of moves found so far at each level.

The use of Gibbs sampling in Monte Carlo Tree Search dates back to the general game player Cadia Player and its MAST playout policy \cite{finnsson2008simulation}.

Monte Carlo Search for combinatorial problems can be much improved using a prior. A prior is a heuristic that is used in playouts to sample in a non uniform way. It favors some moves in the playout according to the heuristic. The use of a bias or the initialization of the weights to produce an initial non uniform policy have been used for multiple difficult problems: the Traveling Salesman Problem with Time Windows \cite{RimmelEvo11,edelkamp2013algorithm} with a distance based heuristic, the Vehicle Routing Problems \cite{edelkamp2016monte,cazenave2021policy} with again a distance based heuristic, the Inverse RNA Folding problem \cite{portela2018unexpectedly} with manually encoded heuristics on pairs of bases, the Pancake problem \cite{Bouzy16} with manually encoded heuristics, the Virtual Network Embedding problem \cite{elkael2022monkey} with a distance based heuristic again. In all these problems the manual prior improves much the performances of Monte Carlo Search.

We propose a method to automatically compute a prior. It uses statistics on solved problems. The method is simple, moreover it does not use computation time during sampling and it is general. It improves much on Monte Carlo Search without a prior for the problems that we tried. It also improves over manually defined priors.

We now give the outline of the paper. The second section describes Monte Carlo Search. The third section explains how to compute the prior. The fourth section gives experimental results for Latin Square Completion (LSC), Kakuro and Inverse RNA Folding.

\section{Monte Carlo Search}

This section presents the GNRPA algorithm which is a generalization of the NRPA algorithm to the use of a prior.

The Nested Rollout Policy Adaptation (NRPA) \cite{Rosin2011} algorithm is an effective combination of NMCS and the online learning of a playout policy. NRPA holds world records for Morpion Solitaire and crosswords puzzles. 

In NRPA/GNRPA each move is associated to a weight stored in an array called the policy. The goal of these two algorithms is to learn these weights using the best sequences of moves found during the  search. The weights are used in the softmax function to produce a playout policy that generates good sequences of moves. 

NRPA/GNRPA use nested search. In NRPA/GNRPA, each level takes a policy as input and returns a sequence and its associated score. At any level $>$ 0, the algorithm makes numerous recursive calls to the lower level, adapting the policy each time with the best sequence of moves to date. The changes made to the policy do not affect the policy in higher levels. At level 0, NRPA/GNRPA return the sequence obtained by the playout function as well as its associated score.

The playout function sequentially constructs a random solution biased by the weights of the moves until it reaches a terminal state. At each step, the function performs Gibbs sampling, choosing the actions with a probability given by the softmax function.

Let $w_{m}$ be the weight associated to a move $m$ in the policy. In NRPA, the probability of choosing move $m$ is defined by: 

$$ p_{m} = \frac{e^{w_{m}}}{\sum_k{e^{w_{k}}}} $$

where $k$ goes through the set of possible moves, including $m$.

GNRPA \cite{Cazenave2020GNRPA} generalizes the way the probability is calculated using a bias $\beta_{m}$. The probability of choosing move $m$ becomes: 
$$ p_{m} = \frac{e^{w_{m}+\beta_{m}}}{\sum_k{e^{w_{k}+\beta_{k}}}} $$

By taking $\beta_{m} = \beta_{k} = 0$, we find the formula for NRPA again. 

In NRPA it is possible to initialize the weights according to a heuristic relevant to the problem to solve. In GNRPA, the policy initialization is replaced by the bias. It is sometimes more practical to use $\beta_{k} $ biases than to initialize the weights as the codes for the moves can be different from the codes of the biases. The method we propose could also be applied without modification to NRPA with initialization of the weights by initializing the weight of move $m$ with $\beta_{m}$ the first time the weight is used.

The algorithm to perform playouts in GNRPA is given in algorithm \ref{PLAYOUT}. The main GNRPA algorithm is given in algorithm \ref{GNRPA}. GNRPA calls the adapt algorithm to modify the policy weights so as to reinforce the best sequence of the current level. The policy is passed by reference to the adapt algorithm which is given in algorithm \ref{ADAPT}.

The principle of the adapt function is to increase the weights of the moves of the best sequence of the level and to decrease the weights of all possible moves by an amount proportional to their probabilities of being played. $\delta_{bm} = 0$ when $b \neq m$ and $\delta_{bm} = 1$ when $b = m$.

\begin{algorithm}
\begin{algorithmic}[1]
\STATE{playout ($policy$)}
\begin{ALC@g}
\STATE{$state \leftarrow root$}
\WHILE{true}
\IF{terminal($state$)}
\RETURN{(score ($state$), $sequence(state)$)}
\ENDIF
\STATE{$z$ $\leftarrow$ 0}
\FOR{$m \in$ possible moves for $state$}
\STATE{$o [m] \leftarrow e^{policy[code(m)] + \beta_m}$}
\STATE{$z \leftarrow z + o [m]$}
\ENDFOR
\STATE{choose a $move$ with probability $\frac{o [move]}{z}$}
\STATE{play ($state$, $move$)}
\ENDWHILE
\end{ALC@g}
\end{algorithmic}
\caption{\label{PLAYOUT}The playout algorithm}
\end{algorithm}

\begin{algorithm}
\begin{algorithmic}[1]
\STATE{adapt ($policy$, $sequence$)}
\begin{ALC@g}
\STATE{$polp \leftarrow policy$}
\STATE{$state \leftarrow root$}
\FOR{$b \in sequence$}
\STATE{$z \leftarrow 0$}
\FOR{$m \in$ possible moves for $state$}
\STATE{$o [m] \leftarrow e^{policy[code(m)] + \beta_m}$}
\STATE{$z \leftarrow z + o [m]$}
\ENDFOR
\FOR{$m \in$ possible moves for $state$}
\STATE{$p_m \leftarrow \frac{o[m]}{z}$}
\STATE{$polp [code(m)] \leftarrow polp [code(m)] - \alpha(p_m - \delta_{bm})$}
\ENDFOR
\STATE{play ($state$, $b$)}
\ENDFOR
\STATE{$policy \leftarrow polp$}
\end{ALC@g}
\end{algorithmic}
\caption{\label{ADAPT}The adapt algorithm}
\end{algorithm}

\begin{algorithm}
\begin{algorithmic}[1]
\STATE{GNRPA ($level$, $policy$)}
\begin{ALC@g}
\IF{level == 0}
\RETURN{playout ($policy$)}
\ELSE
\STATE{$bestScore$ $\leftarrow$ $-\infty$}
\FOR{N iterations}
\STATE{(score,new) $\leftarrow$ GNRPA($level-1$, $policy$)}
\IF{score $\geq$ bestScore}
\STATE{bestScore $\leftarrow$ score}
\STATE{seq $\leftarrow$ new}
\ENDIF
\STATE{adapt ($policy$, $seq$)}
\ENDFOR
\RETURN{(bestScore, seq)}
\ENDIF
\end{ALC@g}
\end{algorithmic}
\caption{\label{GNRPA}The GNRPA algorithm.}
\end{algorithm}

\section{Learning a Prior}

This section presents the computation of the prior. The principle underlying the prior is to compute the frequency each move has been the move solving a problem. In order to compute it we generate many solved problems associated to their solutions, e.g. the sequence of moves that solves the problem from the starting state. It is usually hard for combinatorial problems to find a solution. However in some problems it is easy to generate problems associated to their solutions. The three problems we experimented with have this property that it is easy to generate problems and their associated solutions.

The principle for learning the prior is to replay the solution and to update the count for each possible move of each possible state of the solution. We also update the count of the moves that are part of the solution. We can then calculate for each move the frequency it has been the solution move, this is the number of times it has been in a solution divided by the number of times it has been a possible move.

\begin{algorithm}
\begin{algorithmic}[1]
\STATE{Replay ($state$, $sequence$)}
\begin{ALC@g}
\FOR{$b \in sequence$}
\STATE{$count [code(b)] \leftarrow count [code(b)] + 1$}
\FOR{$m \in$ possible moves for $state$}
\STATE{$nb [code(m)] \leftarrow nb [code(m)] + 1$}
\ENDFOR
\STATE{play ($state$, $b$)}
\ENDFOR
\end{ALC@g}
\end{algorithmic}
\caption{\label{REPLAY}The Replay algorithm}
\end{algorithm}

Algorithm \ref{REPLAY} details how to compute the $count$ and $nb$ arrays given an initial state and the solution to the problem given as a sequence of moves. The $nb$ array memorizes the number of times a move has been possible and the $count$ array memorizes how many times it was part of a solution. The Replay function is called for each solved problem of the training dataset.

We then define the bias $\beta_m$ as:

$$ \beta_m = \tau * log (\frac{count[code(m)]}{nb[code(m)]})$$

where $\tau$ is called the temperature of the bias.

The default sampling policy with a prior plays a move $m$ with probability:

$$p_m = \frac{e^{\beta_m}}{\Sigma_{k} e^{\beta_k}}$$

\section{Experimental Results}

This section details the computation of the prior for three difficult combinatorial problems: Latin Square Completion, Kakuro and Inverse RNA Folding. It also compares sampling with the computed prior to sampling without a prior. It also compares NRPA to GNRPA with the computed prior.

\begin{table*}
  \centering
  \caption{Number of LSC problems of size 20 in the transition phase solved by different algorithms out of 100 problems. The number of playouts ranges from 1,024 playouts to 131,072 playouts. The temperature of the Dual prior is set to $\tau = 4$. Sampling with the Dual prior solves more problems than uniform sampling. GNRPA with the Dual prior is better than NRPA and sampling.}
  \label{LSC}
  \begin{tabular}{lrrrrrrrrrrrrrr}
  Algorithm & 1,024 & 2,048 & 4,096 & 8,192 & 16,384 & 32,768 & 65,536 & 131,072 \\
 ~~~~~~~~~~ &  ~~~~~~~~~~~~ &  ~~~~~~~~~~~~ &   ~~~~~~~~~~~~ & ~~~~~~~~~~~~ & ~~~~~~~~~~~~ & ~~~~~~~~~~~~ & ~~~~~~~~~~~~ & ~~~~~~~~~~~~ & ~~~~~~~~~~~~ & ~~~~~~~~~~~~ \\
Sampling                  & 2 & 5 & 10 & 16 & 26 & 36 & 49 & 61 \\
Sampling Dual prior       & 12 & 24 & 34 & 48 & 70 & 80 & 89 & 95 \\
NRPA                      & 8 & 16 & 25 & 35 & 48 & 61 & 70 & 80 \\
GNRPA Dual prior          & 26 & 39 & 54 & 67 & 83 & 91 & 95 & 98 \\
 \end{tabular}
\end{table*}

\begin{figure}
    \centering
    \includegraphics[width=9cm]{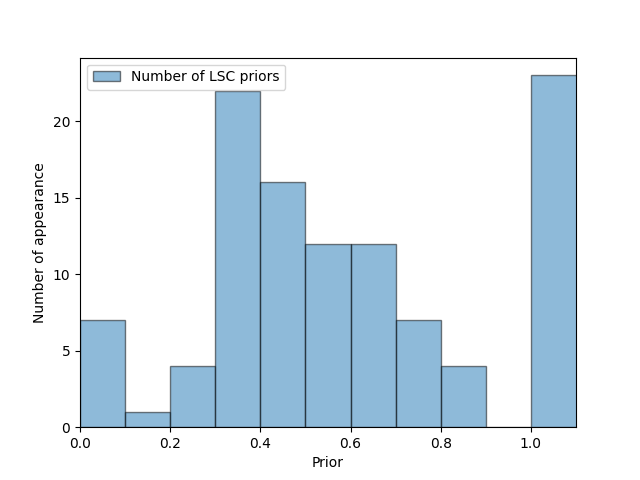}
    \caption{The distribution of the priors for LSC. The priors associated to codes that have never been seen during replay (e.g. nb [code] = 0) have been removed.}
    \label{distribution.latin}
\end{figure}

\subsection{Latin Square Completion}

A Latin Square of order $n$ is a $n \times n$ grid filled with numbers from 1 to $n$ such that the same number does not appear more than once in each row and each column. A partial Latin Square is a Latin Square with some empty cells. The Latin Square Completion problem (LSC) consists in completing a partial Latin Square so as to form a complete Latin Square. Latin Square Completion is a NP-complete problem \cite{colbourn1984complexity}.

\begin{figure}
    \centering
    \includegraphics[width=9cm]{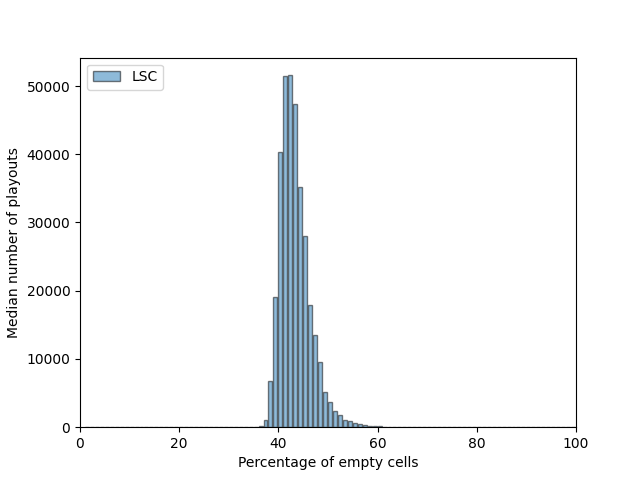}
    \caption{The median number of random playouts required to solve LSC instances of size 20 with x\% of empty cells. The phase transition happens at 42\% of empty cells. All further experiments will use Latin Squares of size 20 with 42\% of empty cells. The median for each percentage was calculated solving 1,000 problems.}
    \label{transition}
\end{figure}

The LSC problem has a phase transition. When a grid has a lot of empty cells or only few empty cells, the completion is very easy. When the percentage of empty cells is close to 42\% the problem becomes hard. Figure \ref{transition} gives the median number of random playouts required to solve LSC problems according to their percentage of empty cells. We can observe the peak in number of random playouts at 42\% of empty cells.

It can also be observed in Figure \ref{transition} that generating Latin Squares from the empty grid is extremely easy. The first three random playouts usually generate a valid Latin Square. Therefore generating difficult LSC problems and their associated solution is also extremely easy. First generate a valid Latin Square, memorize it as a solution and then randomly remove 42\% of the cells so as to have a difficult LSC problem associated to its solution.

Here is an example of a difficult LSC problem of size 20 generated with this method:

\begin{scriptsize}
\begin{verbatim}
 3             11 15  4  8 13 17 14  2    12 10  6          
13    18 17 14 16  8 19  7              1 11       20    12 
20        8    17 12  1 19 10  3  6    16 15          14  4 
             3  4           9 14    15 11        5  7 19  1 
      15 18     9     3  4  1  5  2       13    12  8 10    
      17          19    14              3  7     4 16  6 20 
16 13  4 11       10  9 17              7 14     3 15  5    
    2     1          15 18     6 16  5 12 17 20 19 14 13 11 
 9 20        4     3 11 12  8       17  6    18    19       
11 10 20  6 13           5  3  1        9  4 14 18 12     7 
    8    19          14 10        7 13 18     5     3 17 15 
   12 11  5  6       13       19  4 14 10 20     9  1    18 
19 15  9 20 10     5     3    18     4       17    11  2 13 
18    14  9 16  5     6 11    13 17     2     3     4       
   14  6 15     3  4 18 16  2    11           9  7 17    19 
 4  7  5  3    12          19     9 16 20 18    17          
    3    10     1    16    12  7       17       20  5 18    
    6 12    15  2     7 13        5    19  3        9 11  8 
       7 14 17     1 20       15 13          16       12    
   18    12     8     5       16  3 11    19        6    17 
\end{verbatim}
\end{scriptsize}

LSC and related problems appear in a variety of practical applications such as scheduling, optical routing, error correcting codes as well as combinatorial
design \cite{jin2019solving}.

We model the LSC problem as a Constraint Satisfaction Problem. We use Forward Checking associated to channeling constraints. If a value appears only once in a column or in a row it is directly assigned. If it is not the case, the variable with the smallest number of possible values is chosen and a possible value is randomly assigned according to the policy. A state is terminal if the Latin Square is complete or if one of the variables is not assigned and has an empty domain. In this case the score of a playout is the opposite of the number of remaining variables.

The code associated to a move contains the number of times the value is present in the same column and the number of times it is present in the same row. We call the prior associated to this code the Dual prior. Note that it is a very simple code and that it could probably be refined. The bias for GNRPA using this code is:

$$ \beta_m = \tau * log (\frac{count[code(m)]}{nb[code(m)]})$$

Figure \ref{distribution.latin} gives the distribution of the priors for this code and LSC problems of size 20. The priors were computed using 10,000 solved problems generated randomly in the transition phase. We can observe that the priors have varied values.

Table \ref{LSC} gives the evolution of the number of problems solved by different algorithms with doubling numbers of playouts. Sampling with the Dual prior is much better than sampling without the prior. GNRPA with the Dual prior is much better than NRPA. The computation time of the Dual prior during the playouts is negligible.

\begin{figure}
    \centering
    \includegraphics[width=9cm]{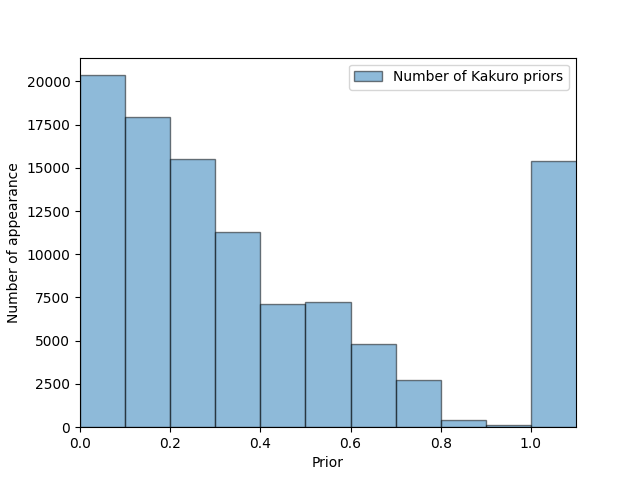}
    \caption{The distribution of the priors for Kakuro. The y-axis gives the number of priors in each range of values. For example there are 15,410 priors that have the value 1.0 and 20,353 priors that have a value between 0.0 and 0.1. The priors associated to codes that have never been seen during replay (e.g. nb [code] = 0) have been removed. We can observe the peak at 0.0 which mainly corresponds to the numbers that are impossible given the row and the column sums. We can also observe the smaller peak at 1.0 which corresponds to the numbers that are forced. Note that apart from these two cases there are many cases where the prior is between 0.0 and 1.0 which does not correspond to a hard constraint.}
    \label{distribution.kakuro}
\end{figure}

\subsection{Kakuro}

\begin{table*}
  \centering
  \caption{Number of Kakuro problems of size 10, with 11 possible values, solved by different algorithms out of 100 problems and for various numbers of playouts. The temperature of the prior is set to $\tau = 4$. Using the prior usually solves the problem in 1 playout.}
  \label{Kakuro}
  \begin{tabular}{lrrrrrrrrrrrrrr}
 Algorithm & 1,024 & 2,048 & 4,096 & 8,192 & 16,384 & 32,768 & 65,536 & 131,072 \\
 ~~~~~~~~~~ &  ~~~~~~~~~~~~ &  ~~~~~~~~~~~~ &   ~~~~~~~~~~~~ & ~~~~~~~~~~~~ & ~~~~~~~~~~~~ & ~~~~~~~~~~~~ & ~~~~~~~~~~~~ & ~~~~~~~~~~~~ & ~~~~~~~~~~~~ & ~~~~~~~~~~~~ \\
Sampling                  & 0 & 0 & 0 & 0 & 0 & 0 & 0 & 0 \\
Sampling Prior            & 100 & 100 & 100 & 100 & 100 & 100 & 100 & 100 \\
NRPA                      & 0 & 0 & 0 & 23 & 35 & 65 & 86 & 98 \\
GNRPA Prior               & 100 & 100 & 100 & 100 & 100 & 100 & 100 & 100 \\
 \end{tabular}
\end{table*}

A Kakuro puzzle is played on a rectangular grid. The objective is to fill numbers into the blank cells, according to the following rules:
\begin{itemize}
\item A sum is associated with every horizontal or vertical sequence of blank cells.
\item Each horizontal (respectively vertical) sequence has a cell left of (respectively above) its first cell, and that cell contains the sum that is associated with the sequence.
\item In each horizontal/vertical sequence of cells, every number may occur at
most once.
\item The sum of the numbers of a sequence must equal the number that is denoted in the corresponding hint.
\end{itemize}

Kakuro is hard \cite{ruepp2010computational}. The most difficult Kakuro problems are the empty problems with only the sum of the columns and of the rows already given \cite{cazenave2009kakuro}.

The generation of a Kakuro problem and its solution is almost as easy as the generation of a LSC problem. First generate a valid square with sampling. A single playout is usually enough. Then calculate the sums for each row and for each column. Then remove all the values and keep the generated valid square as the solution to the problem. Here is an example of a solved Kakuro problem of size 10 with values ranging from 1 to 11 generated with this method:

\begin{verbatim}
        65 60 58 62 59 59 62 60 56 55 
     55  3  5  4  1 10  8  2  9  6  7 
     62  9 11 10  5  3  6  7  1  8  2 
     60  8  2  5 10  9  4 11  3  7  1 
     56  2  4  9  8  1  5  3  7 11  6 
     58  4  7  3  6  2 10  1 11  5  9 
     60  7  1  2  3  8 11  5 10  9  4 
     59  5  3  6 11  4  1  9  8  2 10 
     62 11  9  7  2  6  3 10  5  1  8 
     65  6 10 11  7  5  9  8  2  4  3 
     59 10  8  1  9 11  2  6  4  3  5 
\end{verbatim}

We model Kakuro as a Constraint Satisfaction Problem. We use Forward Checking but we do not use channeling constraints. In a playout we choose the variable with the least number of possible values and we assign a value according to the policy (which is uniform in the case of sampling and which uses the softmax of the biases in the case of the prior policy). When a variable has an empty domain the playout is stopped and the score is returned. The score is the opposite of the number of remaining unassigned variables when the Kakuro is not complete and the number of rows and columns that sum to the hint when all variables are assigned.

The code for a move contains the number of times the value appears in the same row, the number of times it appears in the same column, the remaining sum to reach in the row and the remaining sum to reach in the column.

Figure \ref{distribution.kakuro} gives the numbers of appearance of the Kakuro priors. It was calculated using 10,000 solved problems generated randomly. With the priors equal to 0.0 or 1.0, it rediscovers the hard constraints manually programmed in specialized Kakuro solvers \cite{simonis2008kakuro} that compute the impossible values for a given sum. However our prior is more precise than that since it takes into account the remaining row/column sums as well as the number of appearances of the value in the same row/column. There are many priors different from 0.0 and 1.0 that model something different from the hard constraints and that capture some probabilistic properties of the values to assign.

Table \ref{Kakuro} gives the results for Monte Carlo Search with and without the prior. Using the prior both sampling and GNRPA usually find the solution at the first playout whereas without the prior both sampling and NRPA take much more time.

\subsection{Inverse RNA Folding}

The design of RNA molecules with specific properties is an important topic for health related research. For example, many viruses rely on RNAs to infect and replicate inside a host: this is the case for coronaviruses \cite{madhugiri2018structural} and Dengue viruses. Understanding viral RNAs is essential for the scientific community to develop novel drugs in response to pandemics like COVID-19 \cite{kalvari2021rfam}. 

RNA molecules are long molecules composed of four possible nucleotides. Molecules can be represented as strings composed of the four characters 'A', 'C', 'G', 'U'. For RNA molecules of length N, the size of the state space of possible strings is exponential in N. It can be very large for long molecules. The molecules of the Eterna100 benchmark we use can have hundreds of nucleotides. The sequence of nucleotides folds back on itself to form its secondary structure. It is possible to find in a polynomial time the folded structure of a given sequence. However, the opposite which is to find a sequence that folds into a predefined structure, that is the Inverse RNA Folding problem, is hard \cite{bonnet2020designing}.

The state space is the set of all sequences that are consistent with the secondary structure given as input. The secondary structure is a sequence of characters. The possible characters are '.', '(' and ')'. For each '.' in the input sequence there are four possible characters in the nucleotide sequence: 'A', 'C', 'G' and 'U'. Each '(' character is associated to the ')' character that closes the expression it has opened (e.g. when the same number of '(' and ')' are in between the two). Six pairs of characters are possible to replace the '(' and the corresponding ')': 'CG', 'GC', 'GU', 'UG', 'AU' and 'UA'. When a nucleotide sequence is complete, the ViennaRNA package \cite{lorenz2011viennarna} is used to fold the sequence and verify if it folds into the target structure.

\begin{figure}
    \centering
    \includegraphics[width=5cm]{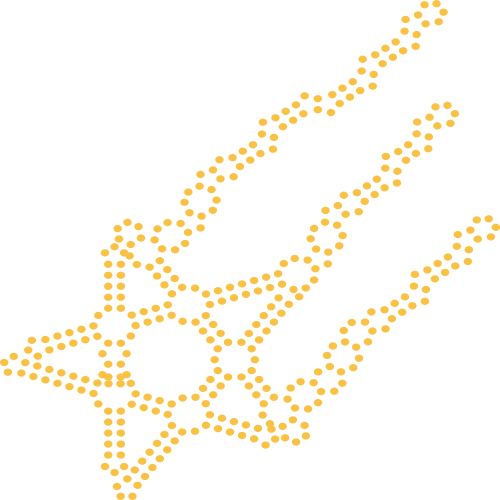}
    \caption{Example of a RNA design puzzle from Eterna100: the Shooting Star. The associated target structure is:\\
(((((((((((((((((...(.(..((.(((.(((.((....).))).)))).).)..))..))))))).(((((..(\\
.(..((.(((.(((((.(((.((((.((....).)))).))))).))).)))).).)..))..)(....)..))))(((\\
((((((...))))).(((((...)))))(.(....).).)))).)))).((((.((((...(((((((...)))))))\\
.(((((...)))))))))(((((((...))))))).(((((...)))))))))((((..(....)(..(..(..(.(.\\
(((.((((.((....).)))).)))).)..)..).)..))))).)))))).
}
    \label{star}
\end{figure}

We evaluate different Monte Carlo Search algorithms on the Eterna100 benchmark which contains 100 RNA secondary structure puzzles of varying degrees of difficulty. A puzzle consists of a given structure under the dot-bracket notation. This notation defines a structure as a sequence of brackets and dots each representing a base. The matching brackets symbolize the paired bases and the dots the unpaired ones. The puzzle is solved when a sequence of the four nucleotides 'A', 'C', 'G' and 'U', folds according to the target structure. In some puzzles, the value of certain bases is imposed. Figure \ref{star} gives an example of a difficult Eterna100 problem.

Human experts have solved the 100 problems of the benchmark. No program has solved all problems. The best score so far for a program is 95/100 by NEMO, NEsted MOnte Carlo RNA Puzzle Solver \cite{portela2018unexpectedly} and by GNRPA using the NEMO prior \cite{Cazenave2020Inverse}.

\begin{figure}
    \centering
    \includegraphics[width=9cm]{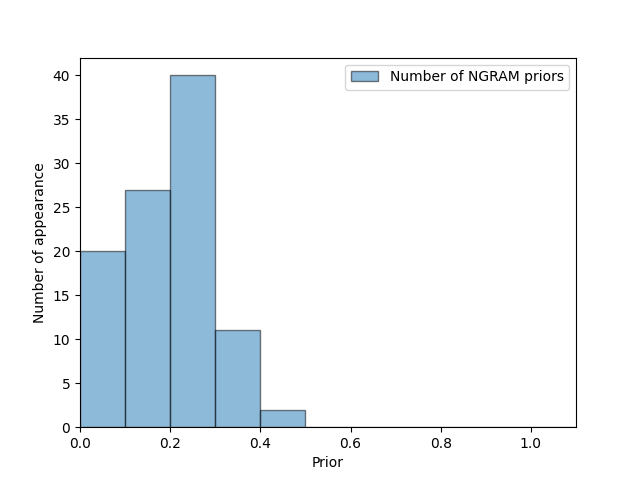}
    \caption{The distribution of the priors for Inverse RNA Folding. The y-axis gives the number of priors in each range of values. There are 6 possible moves for a '(' and 4 possible moves for a '.' in the target structure. This makes 10 possibilities for the previous move in the NGRAM and again 10 possibilities for the current move. Therefore there are 100 different priors. On the contrary of LSC and Kakuro the distribution of the priors is mainly on small values. The smallest prior is equal to 0.010083 and the greatest prior is equal to 0.437825.}
    \label{distribution.ngram}
\end{figure}

\begin{table*}
  \centering
  \caption{Number of Eterna100 problems solved by different algorithms and various search time limits in seconds. GNRPA is much better than NRPA. The NGRAM prior is better than the NEMO prior. The temperature for the NGRAM prior is $\tau = 6$. Sampling with the NGRAM prior is better than sampling with the NEMO prior. Sampling with a prior is much better than uniform sampling.}
  \label{ETERNA}
  \begin{tabular}{lrrrrrrrrrrrrrr}
 Algorithm & 32s & 64s & 128s & 256s & 512s & 1,024s & 2,048s & 4,096s \\
 ~~~~~~~~~~ &  ~~~~~~~~~~~~ &  ~~~~~~~~~~~~ &   ~~~~~~~~~~~~ & ~~~~~~~~~~~~ & ~~~~~~~~~~~~ & ~~~~~~~~~~~~ & ~~~~~~~~~~~~ & ~~~~~~~~~~~~ & ~~~~~~~~~~~~ & ~~~~~~~~~~~~ \\
Sampling                  & 11 & 11 & 11 & 12 & 14 & 16 & 16 & 17 \\
Sampling NEMO prior       & 51 & 55 & 57 & 60 & 61 & 61 & 62 & 64 \\
Sampling NGRAM prior      & 57 & 65 & 68 & 69 & 69 & 69 & 69 & 69 \\
NRPA                      & 28 & 33 & 41 & 48 & 57 & 59 & 61 & 65 \\
GNRPA NEMO prior          & 68 & 69 & 74 & 77 & 78 & 79 & 81 & 81 \\
GNRPA NGRAM prior         & 70 & 75 & 78 & 79 & 80 & 81 & 82 & 85 \\
 \end{tabular}
\end{table*}

\begin{figure}
    \centering
    \includegraphics[width=9cm]{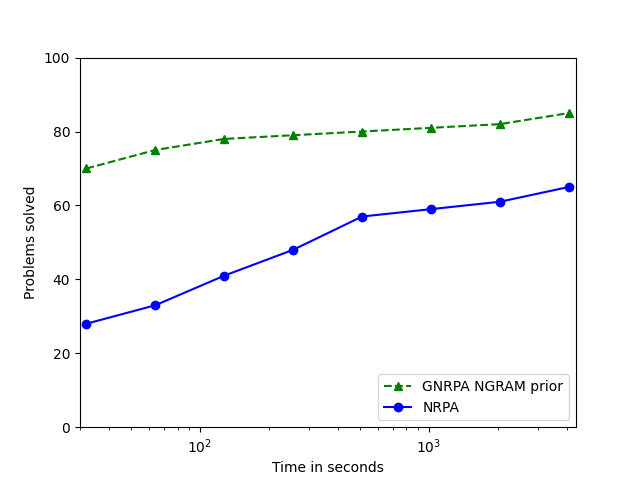}
    \caption{The evolution with the logarithm of the search time of the number of Eterna100 problems solved by NRPA and GNRPA NGRAM prior.}
    \label{plot.ngram}
\end{figure}

\begin{algorithm}
\begin{algorithmic}[1]
\STATE{Replay ($t$, $s$)}
\begin{ALC@g}
\FOR{$i \in [0..len(s)]$}
\FOR{$j \in [0..len(s[i]) - 1]$}
\STATE{$n \leftarrow code (t[i][j],s[i][j],t[i][j+1],s[i][j+1])$}
\STATE{$count[n] \leftarrow count[n] + 1$}
\FOR{$m \in moves(t[i][j+1])$}
\STATE{$n \leftarrow code (t[i][j],s[i][j],t[i][j+1],m)$}
\STATE{$nb[n] \leftarrow nb[n] + 1$}
\ENDFOR
\ENDFOR
\ENDFOR
\end{ALC@g}
\end{algorithmic}
\caption{\label{NGRAM}The algorithm to count the NGRAMs. It takes as arguments the target structures $t$ and the corresponding solutions $s$ as sequences of moves. It counts the number of times two following characters in the target structure and the corresponding two moves happen in the Rfam database. It also counts the number of appearance for all possible moves.}
\end{algorithm}

To compute the prior we use the Rfam database \cite{kalvari2021rfam}. Rfam is the database of non-coding RNA families. We use the 85,232 RNA sequences from Rfam associated to their target folding. 

On the contrary of the hand crafted heuristics of NEMO, the NGRAM prior has been learned on the Rfam database which is separated from the Eterna100 benchmark. The computation of the NGRAM prior on the Rfam database is a more general and simple way to create priors and it is not specific to the Eterna100 benchmark.

Algorithm \ref{NGRAM} gives the function used to compute the NGRAM prior, $t$ is the set of target structures and $s$ is the set of RNA sequences that fold in the target structures. The output of the algorithm are the $count$ and $nb$ arrays that are used to calculate the prior of a move.

The code to calculate the statistics computes $count[code(t_p,m_p,t,k)]$ the number of times an NGRAM coded as $code(t_p,m_p,t,k)$ appears in the solution sequences of the Rfam database. We only compute the NGRAMs of size one, containing $m$ the move to play, $m_p$ the previous move, $t$ the target folding character and $t_p$ the previous target folding character. Figure \ref{distribution.ngram} gives the distribution of the priors.

We define the bias $\beta_m$ as:

$$ \beta_m = \tau * log (\frac{count[code(t_p,m_p,t,m)]}{nb[code(t_p,m_p,t,m)]})$$

The score of a sequence of nucleotide is computed the same way as NEMO \cite{portela2018unexpectedly} using the ViennaRNA package \cite{lorenz2011viennarna}.

Table \ref{ETERNA} gives the evolution of the number of problems solved with time for different Monte Carlo Search algorithms. GNRPA with the NGRAM prior gives the best results. Note that the NEMO prior we used is a subset of the priors used in NEMO. It uses the heuristic functions on the pairs of bases. The pairs of bases heuristics are the main components of the NEMO prior. The same subset of heuristics were already used with GNRPA \cite{Cazenave2020Inverse}, equaling the 95/100 score of NEMO. This score was reached using various optimizations of GNRPA when we use a standard GNRPA in our paper. It explains why we only reach 85 solved problems and why the NEMO prior only reaches 81 solved problems.

Figure \ref{plot.ngram} gives a graphical comparison of NRPA and GNRPA NGRAM prior for the Eterna100 problems. The values for the numbers of solved problems are the same as in the Table \ref{ETERNA}. The time scale is logarithmic.

\section{Conclusion}

Calculating statistics about moves in solved combinatorial problems enables to create a prior for Monte Carlo Search. This prior is easy to compute and has a negligible computation time during sampling. It is a large improvement of Monte Carlo search for three difficult combinatorial problems: Latin Square Completion, Kakuro and Inverse RNA Folding. The method is general and can easily be applied to other difficult combinatorial problems.

As future works, the method could be improved for the combinatorial problems we tried simply using more elaborate codes for the moves. We could bias the policy according to other properties of the moves and of the states than the simple ones we used. The method should also be tried on other difficult combinatorial problems in order to evaluate the gains of using it. The problems we tried are decision problems, it would be interesting to also try optimization problems. The generation of the solved problems would be more time consuming for optimization problems but it would only be done once before the use of the prior in Monte Carlo Search. The sampling time with the prior would be similar to the sampling time without the prior but the scores obtained sampling with the prior could be much better than without the prior.

%\section*{Acknowledgment}

%This work was granted access to the HPC resources of IDRIS under the allocation 2020-AD011011461 and 2020-AD011011714 made by GENCI. This work was supported in part by the French government under management of Agence Nationale de la Recherche as part of the “Investissements d’avenir” program, reference ANR19-P3IA-0001 (PRAIRIE 3IA Institute).

%\clearpage

\bibliographystyle{plain}
\bibliography{main}

\begin{thebibliography}{10}

\bibitem{bonnet2020designing}
Edouard Bonnet, Pawel Rzazewski, and Florian Sikora.
\newblock Designing {RNA} secondary structures is hard.
\newblock {\em Journal of Computational Biology}, 27(3), 2020.

\bibitem{Bouzy13}
Bruno Bouzy.
\newblock Monte-carlo fork search for cooperative path-finding.
\newblock In {\em {Computer Games Workshop} at {IJCAI}}, pages 1--15, 2013.

\bibitem{Bouzy16}
Bruno Bouzy.
\newblock Burnt pancake problem: New lower bounds on the diameter and new
  experimental optimality ratios.
\newblock In {\em {SOCS}}, pages 119--120, 2016.

\bibitem{BrownePWLCRTPSC2012}
Cameron Browne, Edward Powley, Daniel Whitehouse, Simon Lucas, Peter Cowling,
  Philipp Rohlfshagen, Stephen Tavener, Diego Perez, Spyridon Samothrakis, and
  Simon Colton.
\newblock A survey of {M}onte {C}arlo tree search methods.
\newblock {\em {IEEE} Transactions on Computational Intelligence and {AI} in
  Games}, 4(1):1--43, March 2012.

\bibitem{cazenave2009kakuro}
Tristan Cazenave.
\newblock {Monte-Carlo Kakuro}.
\newblock In H.~Jaap van~den Herik and Pieter Spronck, editors, {\em Advances
  in Computer Games, 12th International Conference, {ACG} 2009, Pamplona,
  Spain, May 11-13, 2009. Revised Papers}, volume 6048 of {\em Lecture Notes in
  Computer Science}, pages 45--54. Springer, 2009.

\bibitem{CazenaveIJCAI09}
Tristan Cazenave.
\newblock {Nested Monte-Carlo Search}.
\newblock In Craig Boutilier, editor, {\em IJCAI}, pages 456--461, 2009.

\bibitem{Cazenave2020GNRPA}
Tristan Cazenave.
\newblock Generalized nested rollout policy adaptation.
\newblock In {\em Monte Carlo Search at IJCAI}, 2020.

\bibitem{Cazenave2020Inverse}
Tristan Cazenave and Thomas Fournier.
\newblock Monte {C}arlo inverse folding.
\newblock In {\em Monte Carlo Search at IJCAI}, 2020.

\bibitem{cazenave2021policy}
Tristan Cazenave, Jean-Yves Lucas, Thomas Triboulet, and Hyoseok Kim.
\newblock Policy adaptation for vehicle routing.
\newblock {\em Ai Communications}, 34(1):21--35, 2021.

\bibitem{CazenaveSST16}
Tristan Cazenave, Abdallah Saffidine, Michael~John Schofield, and Michael
  Thielscher.
\newblock Nested monte carlo search for two-player games.
\newblock In {\em {AAAI}}, pages 687--693, 2016.

\bibitem{cazenave2012tsptw}
Tristan Cazenave and Fabien Teytaud.
\newblock Application of the nested rollout policy adaptation algorithm to the
  traveling salesman problem with time windows.
\newblock In {\em Learning and Intelligent Optimization - 6th International
  Conference, {LION} 6}, pages 42--54, 2012.

\bibitem{colbourn1984complexity}
Charles~J. Colbourn.
\newblock The complexity of completing partial latin squares.
\newblock {\em Discrete Applied Mathematics}, 8(1):25--30, 1984.

\bibitem{edelkamp2013algorithm}
Stefan Edelkamp, Max Gath, Tristan Cazenave, and Fabien Teytaud.
\newblock Algorithm and knowledge engineering for the tsptw problem.
\newblock In {\em Computational Intelligence in Scheduling (SCIS), 2013 IEEE
  Symposium on}, pages 44--51. IEEE, 2013.

\bibitem{edelkamp2016monte}
Stefan Edelkamp, Max Gath, Christoph Greulich, Malte Humann, Otthein Herzog,
  and Michael Lawo.
\newblock Monte-{C}arlo tree search for logistics.
\newblock In {\em Commercial Transport}, pages 427--440. Springer International
  Publishing, 2016.

\bibitem{edelkamp2014monte}
Stefan Edelkamp, Max Gath, and Moritz Rohde.
\newblock Monte-{C}arlo tree search for 3d packing with object orientation.
\newblock In {\em KI 2014: Advances in Artificial Intelligence}, pages
  285--296. Springer International Publishing, 2014.

\bibitem{edelkamp2014solving}
Stefan Edelkamp and Christoph Greulich.
\newblock Solving physical traveling salesman problems with policy adaptation.
\newblock In {\em Computational Intelligence and Games (CIG), 2014 IEEE
  Conference on}, pages 1--8. IEEE, 2014.

\bibitem{edelkamp2015monte}
Stefan Edelkamp and Zhihao Tang.
\newblock Monte-{C}arlo tree search for the multiple sequence alignment
  problem.
\newblock In {\em Proceedings of the Eighth Annual Symposium on Combinatorial
  Search, {SOCS} 2015}, pages 9--17. {AAAI} Press, 2015.

\bibitem{elkael2022monkey}
Maxime Elkael, Massinissa~Ait Aba, Andrea Araldo, Hind Castel-Taleb, and Badii
  Jouaber.
\newblock Monkey business: Reinforcement learning meets neighborhood search for
  virtual network embedding.
\newblock {\em Computer Networks}, 216:109204, 2022.

\bibitem{finnsson2008simulation}
Hilmar Finnsson and Yngvi Bj{\"o}rnsson.
\newblock Simulation-based approach to general game playing.
\newblock In {\em AAAI}, volume~8, pages 259--264, 2008.

\bibitem{jin2019solving}
Yan Jin and Jin-Kao Hao.
\newblock Solving the latin square completion problem by memetic graph
  coloring.
\newblock {\em IEEE Transactions on Evolutionary Computation},
  23(6):1015--1028, 2019.

\bibitem{kalvari2021rfam}
Ioanna Kalvari, Eric~P Nawrocki, Nancy Ontiveros-Palacios, Joanna Argasinska,
  Kevin Lamkiewicz, Manja Marz, Sam Griffiths-Jones, Claire Toffano-Nioche,
  Daniel Gautheret, Zasha Weinberg, et~al.
\newblock Rfam 14: expanded coverage of metagenomic, viral and microrna
  families.
\newblock {\em Nucleic Acids Research}, 49(D1):D192--D200, 2021.

\bibitem{lorenz2011viennarna}
Ronny Lorenz, Stephan~H Bernhart, Christian H{\"o}ner~zu Siederdissen, Hakim
  Tafer, Christoph Flamm, Peter~F Stadler, and Ivo~L Hofacker.
\newblock Viennarna package 2.0.
\newblock {\em Algorithms for molecular biology}, 6:1--14, 2011.

\bibitem{madhugiri2018structural}
Ramakanth Madhugiri, Nadja Karl, Daniel Petersen, Kevin Lamkiewicz, Markus
  Fricke, Ulrike Wend, Robina Scheuer, Manja Marz, and John Ziebuhr.
\newblock Structural and functional conservation of cis-acting rna elements in
  coronavirus 5'-terminal genome regions.
\newblock {\em Virology}, 517:44--55, 2018.

\bibitem{Mehat2010}
Jean M{\'e}hat and Tristan Cazenave.
\newblock Combining {UCT} and {Nested Monte Carlo Search} for single-player
  general game playing.
\newblock {\em {IEEE} Transactions on Computational Intelligence and {AI} in
  Games}, 2(4):271--277, 2010.

\bibitem{portela2018unexpectedly}
Fernando Portela.
\newblock An unexpectedly effective {M}onte {C}arlo technique for the {RNA}
  inverse folding problem.
\newblock {\em BioRxiv}, page 345587, 2018.

\bibitem{PouldingF14}
Simon~M. Poulding and Robert Feldt.
\newblock Generating structured test data with specific properties using nested
  {M}onte-{C}arlo search.
\newblock In {\em {GECCO}}, pages 1279--1286, 2014.

\bibitem{PouldingF15}
Simon~M. Poulding and Robert Feldt.
\newblock Heuristic model checking using a {M}onte-{C}arlo tree search
  algorithm.
\newblock In {\em {GECCO}}, pages 1359--1366, 2015.

\bibitem{RimmelEvo11}
Arpad Rimmel, Fabien Teytaud, and Tristan Cazenave.
\newblock Optimization of the {Nested Monte-Carlo} algorithm on the traveling
  salesman problem with time windows.
\newblock In {\em EvoApplications}, volume 6625 of {\em LNCS}, pages 501--510.
  Springer, 2011.

\bibitem{Rosin2011}
Christopher~D. Rosin.
\newblock Nested rollout policy adaptation for {Monte Carlo Tree Search}.
\newblock In {\em {IJCAI} 2011, Proceedings of the 22nd International Joint
  Conference on Artificial Intelligence}, pages 649--654, 2011.

\bibitem{roucairol2023retrosynthesis}
Milo Roucairol and Tristan Cazenave.
\newblock Comparing search algorithms on the retrosynthesis problem.
\newblock In {\em AI to Accelerate Science and Engineering at AAAI 2023}. 2023.

\bibitem{ruepp2010computational}
Oliver Ruepp and Markus Holzer.
\newblock The computational complexity of the kakuro puzzle, revisited.
\newblock In {\em International Conference on Fun with Algorithms}, pages
  319--330. Springer, 2010.

\bibitem{silver2017mastering}
David Silver, Thomas Hubert, Julian Schrittwieser, Ioannis Antonoglou, Matthew
  Lai, Arthur Guez, Marc Lanctot, Laurent Sifre, Dharshan Kumaran, Thore
  Graepel, Timothy~P. Lillicrap, Karen Simonyan, and Demis Hassabis.
\newblock Mastering chess and shogi by self-play with a general reinforcement
  learning algorithm.
\newblock {\em CoRR}, abs/1712.01815, 2017.

\bibitem{simonis2008kakuro}
Helmut Simonis.
\newblock Kakuro as a constraint problem.
\newblock {\em Proc. seventh Int. Works. on Constraint Modelling and
  Reformulation}, 2008.

\end{thebibliography}

\end{document}